%% file: main.tex
\documentclass[sigconf,nonacm]{acmart}

\AtBeginDocument{%
  \providecommand\BibTeX{{\rmfamily B\kern-.05em{\scshape i\kern-.025em b}\kern-.08em
  T\kern-.1667em\lower.7ex\hbox{E}\kern-.125emX}}}

\usepackage{float}
\usepackage{placeins}
\renewcommand\footnotetextcopyrightpermission[1]{}

\begin{document}

\title{TS-RAG: Retrieval Augmented Generation for Time Series Forecasting}

\author{Yixiong Xiao}
\email{xiaoyixiong@baidu.com}
\affiliation{%
  \institution{Baidu, Inc.}
  \city{Beijing}
  \country{China}
}

\author{Congxi Xiao}
\email{xiaocongxi@baidu.com}
\affiliation{%
  \institution{Baidu, Inc.}
  \city{Beijing}
  \country{China}
}

\author{Shuangli Li}
\affiliation{%
  \institution{Baidu, Inc.}
  \city{Beijing}
  \country{China}
}

\author{Jingbo Zhou}
\authornote{Corresponding author.}
\email{jingbozhou@baidu.com}
\affiliation{%
  \institution{Baidu, Inc.}
  \city{Beijing}
  \country{China}
}
\renewcommand{\shortauthors}{Xiao et al.}

\begin{abstract}
While deep learning models, particularly transformer-based architectures, have shown impressive performance in time series forecasting, the application of retrieval-augmented generation (RAG) in this domain remains limited. Since RAG has proven effective in enhancing the capabilities of large language models by incorporating relevant external information, retrieving similar time series sequences as references might also improve accuracy in time series forecasting tasks. However, most time series models are constrained by limited training data, smaller parameter scales, and a lack of the extensive generative capabilities found in large language models. Simply concatenating reference sequences into the prompt, as done in language models, may not yield the expected results. To address these challenges, we propose a novel approach, TS-RAG, which leverages RAG to enhance forecasting performance. The framework introduces specially designed reference tokens to effectively fuse information from the input sequence with that from retrieved similar sequences, enabling a more robust capture of complex temporal dynamics. Experimental results demonstrate that TS-RAG achieves consistent state-of-the-art performance across several real-world forecasting benchmarks.
\end{abstract}

\keywords{time series forecasting, retrieval augmented generation, RAG}

\begin{teaserfigure}
  \centering
  \includegraphics[width=\textwidth]{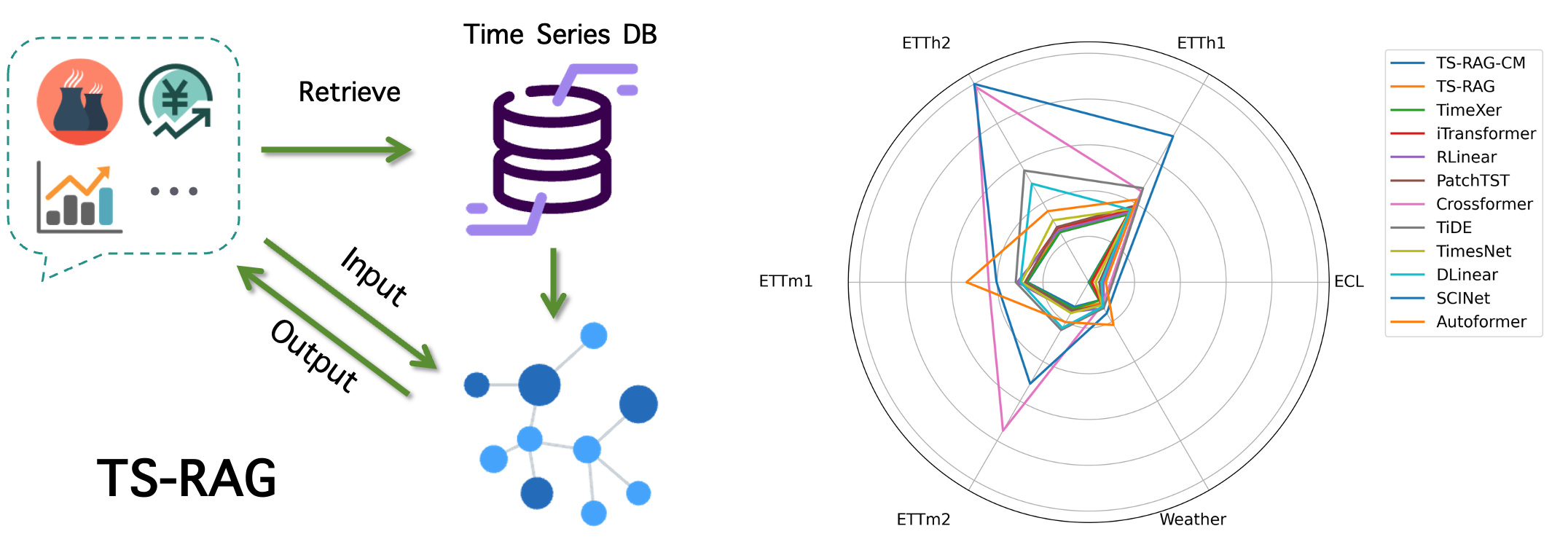}
  \caption{Left: schematic overview of TS-RAG. Right: model performance comparison on multiple forecasting benchmarks.}
  \Description{A TS-RAG overview showing retrieval and forecasting components, alongside a radar chart comparing forecasting models on ECL, ETTh1, ETTh2, ETTm1, ETTm2, and Weather datasets.}
  \label{fig:teaser}
\end{teaserfigure}

\maketitle

\input{section/introduction}
\input{section/relatedwork}
\input{section/methodology}

\input{section/experiment}
\input{section/conclusion}

\bibliographystyle{ACM-Reference-Format}
\bibliography{references}

\end{document}

%% file: section/introduction.tex
\section{Introduction}

Time series forecasting plays a crucial role in numerous real-world applications, including finance, healthcare, energy, and climate science. The ability to accurately predict future values based on historical data is essential for decision-making processes across these domains. In recent years, deep learning models, particularly transformer-based architectures, have demonstrated remarkable performance in time series forecasting due to their ability to model long-range dependencies and capture complex temporal patterns \cite{vaswani2017attention}. These models excel at handling sequential data by using self-attention mechanisms that allow them to focus on relevant parts of the input sequence, even over long time horizons. Despite these advances, most existing models primarily focus on enhancing model training to improve accuracy, such as incorporating multi-scale mixing \cite{zhou2021multi} and time imaging techniques \cite{li2020time}. These approaches aim to capture the intricate relationships and dependencies across different time scales within the time series. However, research into how to effectively leverage retrieval-augmented mechanisms to enhance forecasting accuracy is relatively scarce.

Retrieval-Augmented Generation (RAG) has emerged as an effective paradigm in natural language processing (NLP), where it enhances large language models by retrieving and incorporating relevant external information during inference. This approach has been shown to improve the accuracy and generalization of LLMs, particularly in knowledge-intensive tasks, by enabling the model to access external sources of information beyond its training data. RAG leverages the ability of retrieval systems to pull in pertinent data from large corpora, ensuring that the model’s outputs are informed by relevant examples, improving its robustness and flexibility. In the context of time series forecasting, time series data often exhibits highly intricate patterns, such as seasonality, trends, and volatility. RAG can improve performance by retrieving similar historical sequences that share key characteristics with the current input, allowing the model to pay closer attention to the intricate patterns in these similar sequences, thereby improving prediction accuracy. From this perspective, extending RAG from NLP to time series forecasting thus presents a promising approach.

However, incorporating RAG into time series models faces several challenges. While a straightforward adaptation might involve retrieving similar historical sequences and concatenating them with the input sequence, as is done in large language models (LLMs), this approach may not be directly effective in time series forecasting because it relies on the generalization ability of large language models. Despite some efforts to develop foundational time series models, most are still limited by smaller model sizes and insufficient training data, lacking the generative capabilities of LLMs. As a result, simply appending retrieved sequences to the input does not facilitate effective information fusion, since time series models generally lack the extensive generative power of LLMs and require specialized methods to integrate external information meaningfully. Another key challenge is the efficiency of similarity retrieval, particularly when using traditional methods like Dynamic Time Warping (DTW). These methods can be computationally expensive, requiring substantial time to identify similar sequences from large datasets. As a result, even if retrieval improves forecasting performance, the time required for inference can be prohibitive, limiting the scalability and real-time applicability of such models.

To address these challenges, we propose TS-RAG, a novel framework that leverages retrieval-augmented generation for time series forecasting. TS-RAG introduces specially designed reference tokens that effectively integrate information from both the input sequence and retrieved similar sequences. This mechanism not only helps fuse complex temporal dependencies from historical data but also ensures that external information is incorporated meaningfully. Moreover, by employing vector search techniques for efficient similarity retrieval, TS-RAG significantly reduces the computational burden associated with traditional methods like Dynamic Time Warping (DTW). As a result, TS-RAG enables faster inference times without compromising on predictive accuracy, enhancing generalization across diverse forecasting tasks.

We evaluate TS-RAG on multiple real-world time series forecasting benchmarks, demonstrating that it consistently outperforms existing state-of-the-art methods. Our results highlight the potential of retrieval-augmented generation in advancing time series forecasting by enabling models to leverage external information in a meaningful and structured way. This work contributes to bridging the gap between retrieval-based augmentation techniques and time series forecasting, opening new avenues for improving predictive performance with additional references.

%% file: section/relatedwork.tex
\section{Related Work}

\subsection{Time Series Forecasting}

Deep learning has been widely applied in time series forecasting, primarily utilizing CNNs and RNNs. Convolutional Neural Networks (CNNs) extract localized temporal features using convolutional filters, making them effective in capturing short-term dependencies in time series data~\cite{bai2018cnn, borovykh2017cnn}. Recurrent Neural Networks (RNNs) and their variants, such as Long Short-Term Memory (LSTM) networks and Gated Recurrent Units (GRUs), have been extensively used for sequence modeling due to their ability to capture long-term dependencies~\cite{hewamalage2021lstm}. However, RNN-based models suffer from limitations such as vanishing gradients and sequential computation inefficiencies, making them less scalable for long-horizon forecasting tasks.  

The introduction of Transformer-based architectures~\cite{vaswani2017attention} revolutionized time series forecasting by providing a mechanism for capturing long-range dependencies without the need for sequential processing. Unlike RNNs, Transformers use self-attention mechanisms to model dependencies across arbitrary time steps, significantly improving efficiency and scalability. Several Transformer variants have been developed specifically for time series forecasting. LogTrans~\cite{li2019logtrans} introduced convolutional self-attention to improve efficiency. Informer~\cite{zhou2021informer} proposed the probSparse self-attention mechanism, reducing computational complexity while preserving predictive accuracy. Autoformer~\cite{wu2021autoformer} incorporated seasonal-trend decomposition, making the model more robust to non-stationary patterns. FedFormer~\cite{zhou2022fedformer} extended this concept by using frequency-domain attention to improve long-range forecasting. Despite their effectiveness, Transformer-based models still face scalability challenges, particularly in handling large-scale datasets and high-dimensional multivariate forecasting. Recent studies suggest that even simple linear models can achieve competitive results~\cite{zeng2023linear}, leading to the development of lightweight architectures such as MLP-based forecasting models~\cite{chen2023tsmixer, das2023tide, wang2024timemixer}. These models use time series decomposition techniques and multi-periodicity analysis to efficiently extract temporal features while reducing computational overhead.  

The recent emergence of Time Series Foundation Models (TSFMs) has further advanced time series forecasting. Unlike traditional models trained on domain-specific datasets, TSFMs leverage large-scale, multi-domain pretraining to achieve zero-shot forecasting capabilities. TimeGPT-1~\cite{garza2023timegpt} was the first closed-source TSFM, demonstrating the potential of large pre-trained models for general time series prediction. ForecastPFN~\cite{dooley2024forecastpfn} achieved strong zero-shot forecasting performance by pretraining on synthetic time series data. TimesFM~\cite{das2024timesfm} introduced patch-based decoder architectures, enhancing computational efficiency for long-horizon forecasting. To further improve generalization, recent works have focused on multi-source training datasets. MOIRAI~\cite{woo2024moirai} constructed LOTSA, a large open-source time series dataset, and trained a masked encoder-based foundation model that achieved competitive or superior performance to full-shot models. Tiny Time Mixers (TTMs)~\cite{ekambaram2024ttm} adopted a lightweight mixing-based architecture, demonstrating strong zero-shot forecasting capabilities. Although TSFMs significantly improve generalization across domains, they still rely heavily on implicit knowledge learned during pretraining, limiting their ability to adapt to rare, non-stationary, or previously unseen time series patterns. To address this challenge, recent research has introduced retrieval-augmented methods, which explicitly incorporate external knowledge into the forecasting process.  

\subsection{Retrieval-Augmented Generation (RAG)}

Retrieval-Augmented Generation (RAG) has been widely explored in natural language processing (NLP) as a method to enhance model capabilities by retrieving external knowledge during inference~\cite{lewis2020rag, guu2020retrieval}. Traditional large language models (LLMs) often suffer from knowledge cutoffs and hallucinations, as they rely solely on parametric memory. RAG mitigates these limitations by dynamically retrieving relevant information from large-scale corpora, ensuring more accurate and contextually relevant responses~\cite{karpukhin2020dpr}. In text generation, RAG has been applied to knowledge-grounded tasks such as question answering, summarization, and fact verification, significantly improving zero-shot and few-shot performance. Furthermore, in dialogue systems, RAG enables more personalized and context-aware responses by retrieving user-specific information or relevant conversational history~\cite{huang2023lapdog}. Compared to fully parametric models, retrieval-augmented approaches allow for more modular and interpretable knowledge integration, reducing the need for continual fine-tuning while improving adaptability across different domains.

Beyond NLP, retrieval-augmented techniques have been successfully applied in other domains, including computer vision and structured data modeling. In image generation, retrieval-based models enhance quality by selecting relevant visual features, style references, or object compositions from external image databases~\cite{chen2023reimagen}. This allows for more detailed and coherent visual outputs compared to purely generative approaches. In structured data tasks, retrieval-based learning has been explored in various contexts, such as nearest-neighbor retrieval~\cite{zhang2016knn}, kernel-based retrieval~\cite{nader2022kernel}, and prototype-based learning~\cite{arik2020prototype}. These techniques demonstrate that integrating external information improves prediction accuracy and robustness in domains like finance, healthcare, and recommendation systems. More recently, retrieval-augmented deep learning methods have been proposed for tabular data forecasting, where models leverage attention-like retrieval mechanisms to select relevant historical patterns and enhance prediction accuracy~\cite{gorishniy2024retrieval}.

Despite its success in NLP and structured data applications, retrieval-augmented forecasting remains underexplored in time series research. Traditional forecasting models assume that all necessary information is encoded within model weights, limiting adaptability to unseen distributions. Early retrieval-based forecasting methods, such as ReTime~\cite{jing2022retime}, introduced relational retrieval mechanisms to improve forecasting and imputation for incomplete sequences. More recently, RAFT (Retrieval-Augmented Forecasting of Time-Series)~\cite{tire2024retrievalaugmentedtimeseries} proposed an explicit retrieval module that selects historical patterns similar to the input sequence from the training dataset. By incorporating retrieved patterns alongside learned representations, RAFT enhances forecasting accuracy while maintaining model efficiency. Unlike Transformer-based approaches that rely solely on self-attention to implicitly capture temporal dependencies, RAFT explicitly retrieves and integrates relevant past sequences, improving generalization to diverse time series distributions.

%% file: section/methodology.tex
\section{Methodology}
\begin{figure*}
  \includegraphics[width=\textwidth]{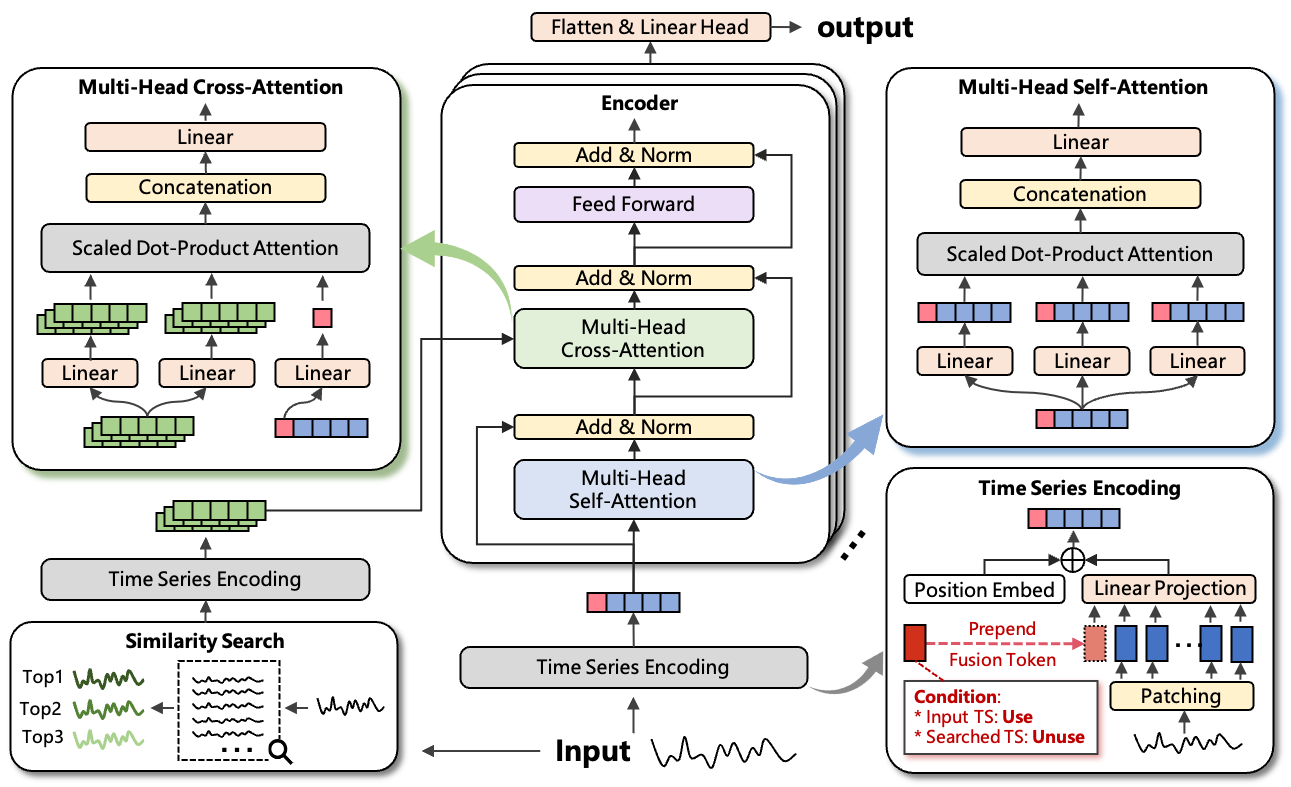}
  \caption{The Framework of TS-RAG. }
  \Description{The TS-RAG architecture combines similarity search, time-series encoding, self-attention, cross-attention, reference fusion tokens, and a forecasting head.}
  \label{fig:framework}
\end{figure*}
\subsection{Problem Definition}

Given a multivariate time series \( \mathbf{x}^{(i)}_{1:L} \in \mathbb{R}^{C \times L} \), where \( C \) denotes the number of channels (or variables) and \( L \) represents the sequence length, the objective is to predict the future values:

\begin{equation}
    \mathbf{x}^{(i)}_{L+1:L+T} \in \mathbb{R}^{C \times T}.
\end{equation}

Traditional forecasting models rely solely on the given sequence \( \mathbf{x}^{(i)}_{1:L} \) to capture temporal dependencies. However, real-world time series often exhibit recurring patterns across different periods, events, or contexts. Retrieval-augmented forecasting enhances predictive performance by integrating relevant past sequences that share similar trends.

We assume access to a historical database \( \mathcal{D} \) containing a collection of past time series segments. Using a similarity-based retrieval function \( \mathcal{R} \), we identify and retrieve the most relevant sequences:

\begin{equation}
    r^{(i)} = \mathcal{R}\bigl(\mathbf{x}^{(i)}, \mathcal{D}\bigr) 
    = \{r^{(i,n)}\}_{n=1}^{N}, \quad r^{(i,n)} \in \mathbb{R}^{C \times L}.
\end{equation}

These retrieved sequences provide additional historical context that can aid forecasting. The forecasting task is thus reformulated as:

\begin{equation}
    \hat{\mathbf{x}}^{(i)}_{L+1:L+T} 
    = f_{\theta}\bigl(\mathbf{x}^{(i)}_{1:L}, r^{(i)}\bigr),
\end{equation}

where \( f_{\theta} \) represents the retrieval-augmented forecasting model. By incorporating these reference sequences, the model gains a broader temporal perspective, allowing it to detect patterns that may not be apparent from the input sequence alone.

\subsection{Embedding Process}

\subsubsection{Input Sequence Embedding}

To effectively process the time series data, each channel \( c \) of \( \mathbf{x}^{(i)}_{1:L} \) is divided into non-overlapping patches of length \(p\):
\begin{equation}
    \mathbf{x}^{(i,c)}_{\text{patch}} 
    = \Bigl( 
      \mathbf{x}^{(i,c)}_{1:p}, \mathbf{x}^{(i,c)}_{p+1:2p}, 
      \dots, 
      \mathbf{x}^{(i,c)}_{L-p+1:L}
    \Bigr) 
    \in \mathbb{R}^{\frac{L}{p} \times p}.
\end{equation}



To incorporate retrieved reference sequences, we follow an idea reminiscent of "global tokens" in multi-modal contexts. Specifically, each retrieved sequence \( r^{(i,n)} \) is associated with a learnable reference token \( \mathbf{v}_{\text{ref}}^{(n)}(c) \). These tokens act as compressed representations of historical patterns and are prepended to the patchified input:

\begin{equation}
    \tilde{\mathbf{x}}^{(i,c)}_{\text{patch}} 
    = \Bigl(
      \mathbf{v}_{\text{ref}}^{(1)}(c), 
      \mathbf{v}_{\text{ref}}^{(2)}(c), 
      \dots, 
      \mathbf{v}_{\text{ref}}^{(N)}(c), 
      \mathbf{x}^{(i,c)}_{\text{patch}}
    \Bigr).
\end{equation}

The final input embedding is computed as:

\begin{equation}
    \mathbf{x}_{\text{embed}}^{(i)} 
    = W_x\, \tilde{\mathbf{x}}^{(i)}_{\text{patch}} + b_x 
    + \mathrm{PE}(\tilde{\mathbf{x}}^{(i)}_{\text{patch}}),
\end{equation}

where \( W_x \) and \( b_x \) are learnable parameters, and \( \mathrm{PE}(\cdot) \) denotes the positional encoding function.

\subsubsection{Reference Sequences Embedding}

Each retrieved sequence \( r^{(i,n)} \) undergoes independent patchification:

\begin{equation}
    \mathbf{r}^{(i,c,n)}_{\text{patch}} 
    = \Bigl( 
      r^{(i,c,n)}_{1:p}, 
      \dots, 
      r^{(i,c,n)}_{L-p+1:L} 
    \Bigr).
\end{equation}


The corresponding embeddings are then computed as:

\begin{equation}
    \mathbf{r}_{\text{embed}}^{(i,n)} 
    = W_x \mathbf{r}^{(i,n)}_{\text{patch}} + b_x 
    + \mathrm{PE}(r^{(i,n)}_{\text{patch}}).
\end{equation}

\subsection{Reference Fusion}

To integrate both the input sequence and retrieved sequences, we utilize self-attention and cross-attention mechanisms:

\begin{enumerate}
    \item \textbf{Self-Attention}: Captures intrinsic temporal dependencies within the input sequence.
    \item \textbf{Cross-Attention}: Aligns retrieved sequences with the input sequence, allowing the model to selectively extract useful information.
\end{enumerate}

\subsubsection{Self-Attention}

Self-attention is applied to the target sequence embedding \( \mathbf{h}_x^{(i)} \):

\begin{equation}
    \mathbf{h}_x^{\prime(i)} 
    = \operatorname{softmax} 
    \Bigl( 
      \frac{\mathbf{h}_x^{(i)} W_Q (\mathbf{h}_x^{(i)} W_K)^\top}{\sqrt{d}} 
    \Bigr) \mathbf{h}_x^{(i)} W_V.
\end{equation}

\subsubsection{Cross-Attention}

Cross-attention is applied to integrate retrieved sequences:

\begin{equation}
    \mathbf{h}_{x,r}^{(i)} 
    = \operatorname{softmax} 
    \Bigl( 
      \frac{\mathbf{h}_x^{(i)} W_Q (\mathbf{h}_r^{(i)} W_K)^\top}{\sqrt{d}} 
    \Bigr) \mathbf{h}_r^{(i)} W_V.
\end{equation}

The final hidden representations are passed through a feed-forward network, and the model outputs the forecast \( \hat{\mathbf{x}}^{(i)}_{L+1:L+T} \), leveraging both self-learned representations and retrieved historical patterns to improve predictive accuracy.

%% file: section/experiment.tex
\section{Experiments}

\noindent \textbf{Dataset} We conduct our experiments on six widely used benchmark datasets for long-term time series forecasting: ETTh1, ETTh2, ETTm1, ETTm2, Electricity, and Weather. These datasets span various domains, including energy consumption, electricity load forecasting, and meteorological measurements, ensuring a comprehensive evaluation of our model's performance. The ETTh1 and ETTh2 datasets contain hourly transformer temperature and power load readings, while ETTm1 and ETTm2 provide minute-level variations of the same. The Electricity dataset records hourly consumption data of 321 customers, making it suitable for demand forecasting tasks. The Weather dataset contains 21 meteorological indicators, including temperature, humidity, wind speed, and pressure. We follow the standard data preprocessing and train-validation-test splits as previous works, including TimeXer~\cite{wang2024timexerempoweringtransformerstime} and iTransformer~\cite{liu_itransformer_2024}, ensuring a fair comparison. The input sequence length is set to 96 time steps, and the model predicts the next 96, 192, 336, or 720 time steps.

\noindent \textbf{Baselines} To evaluate the effectiveness of our approach, we compare it against several state-of-the-art time series forecasting models, including Transformer-based and retrieval-augmented models. The Transformer-based baselines include Informer~\cite{zhou2021informer}, which introduced the probSparse self-attention mechanism to improve efficiency, Autoformer~\cite{wu2021autoformer}, which leveraged seasonal-trend decomposition for better forecasting interpretability, FedFormer~\cite{zhou2022fedformer}, which incorporated frequency-domain attention mechanisms, PatchTST~\cite{nie_time_2023}, which utilized patch-based attention mechanisms for long-range dependencies, and TimeXer~\cite{wang2024timemixer}, which explicitly modeled exogenous variables through patch-wise and variate-wise cross-attention mechanisms. Our model follows the same experimental settings as Timer to ensure a direct comparison between retrieval-based approaches and other forecasting methods.

\noindent \textbf{Implementation Details} All experiments are implemented in PyTorch and conducted on an NVIDIA V100 GPU with 32GB of memory. To ensure consistency across experiments, we adopt the same hyperparameter settings as iTransformer \cite{liu_itransformer_2024}. Our model employs a retrieval-augmented structure, where a retriever selects the most relevant historical patterns based on the input sequence.
We train the model using a sliding window approach, with an input length of 96 and forecasting horizons of 96, 192, 336, and 720. Training is performed using the Adam optimizer with hyperparameters $\beta_1 = 0.9$, $\beta_2 = 0.999$, and a learning rate of 0.001 with cosine decay. Additionally, we train a Temporal Convolutional Network (TCN) to retrieve time series sequences most similar to the input. In our ablation study, we compare retrieval performance using Euclidean distance and Dynamic Time Warping (DTW).
For evaluation, we retrieve the most similar sequences to the input to generate initial forecast results. We also report the impact of varying the number of retrieved sequences in the ablation study. Model performance is measured using Mean Squared Error (MSE) and Mean Absolute Error (MAE), with all reported results averaged over five independent runs to account for experimental variability.
\input{tables/main_result}

\subsection{Main Results}
Table \ref{tab:main_result} presents the results of our model alongside baseline models. A lower MSE or MAE indicates better forecasting performance. TS-RAG-CM and TS-RAG denote models that consider and do not consider channel dependency, respectively. As shown in the table, TS-RAG-CM achieves the best or near-best performance across all six datasets (ECL, ETTh1, ETTh2, ETTm1, ETTm2, Weather), demonstrating superior predictive accuracy. Specifically, it achieves the lowest average MSE (0.310) and MAE (0.348), outperforming all other models. This result validates the effectiveness of retrieval-augmented generation (RAG) in enhancing multivariate time series forecasting performance.

TS-RAG also exhibits competitive results, ranking third overall. Notably, TS-RAG only considers relationships between the reference token, the input sequence within the same channel, and the corresponding reference sequence, without explicitly modeling cross-channel dependencies. Despite this limitation, it still achieves results comparable to iTransformer, which explicitly incorporates channel mixing. This highlights the significant contribution of RAG in a channel-independent setting, underscoring its potential to enhance time series forecasting even without explicit inter-channel interactions.

In contrast, traditional regression-based methods such as RLinear and DLinear, as well as deep learning baselines like PatchTST, Crossformer, TiDE, and TimesNet, generally underperform compared to TS-RAG-CM. This further reinforces the effectiveness of retrieval-augmented approaches in multivariate time series forecasting.

\subsection{Ablation Study}
Compared to traditional time series forecasting models, TS-RAG introduces a retrieval module that brings additional considerations, such as the choice of retrieval method, the number of retrieved sequences, and the integration strategy with the forecasting model. Each of these factors can significantly impact the final predictive performance. Therefore, our ablation study focuses on systematically analyzing and comparing the effects of these components on model performance. 
\begin{figure*}
  \includegraphics[width=\textwidth]{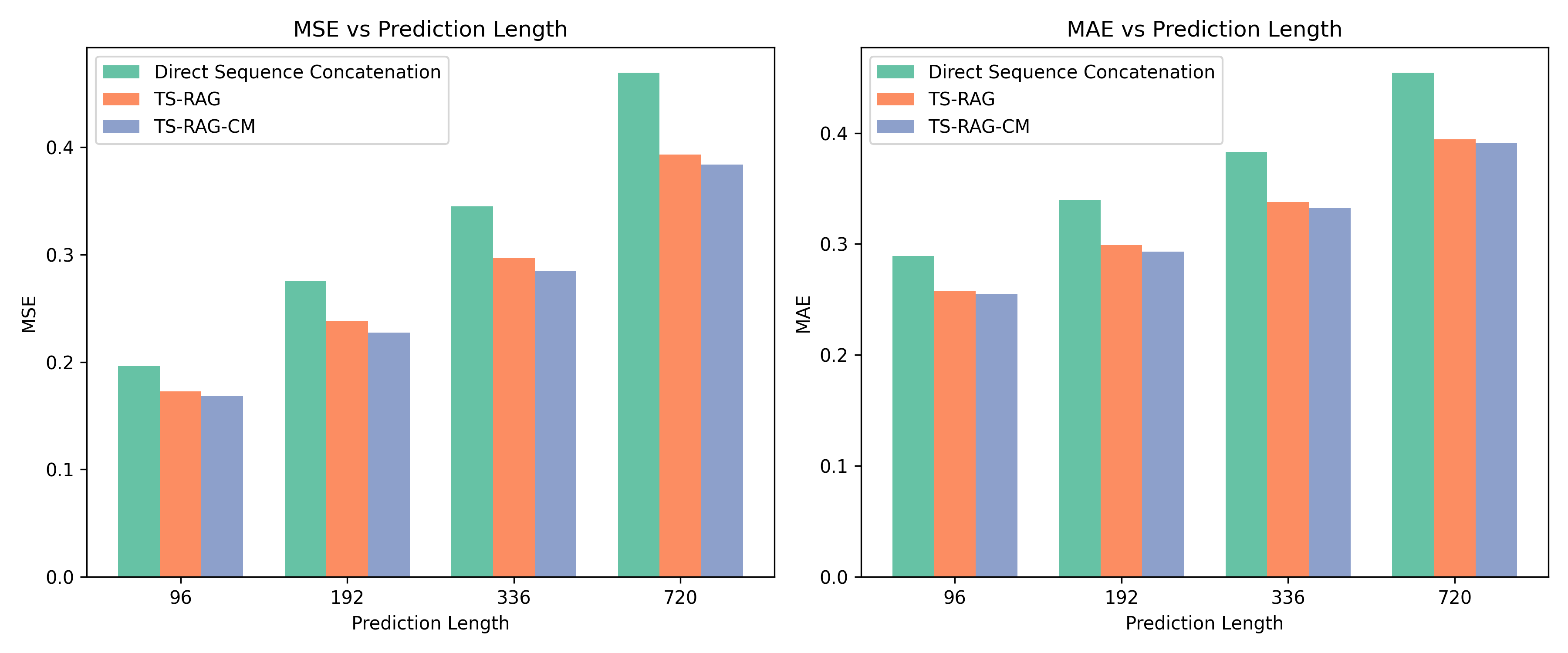}
  \caption{Comparison of TS-RAG-CM, TS-RAG with Direct Sequence Concatenation. }
  \Description{Forecasting error comparison between TS-RAG-CM, TS-RAG, and direct sequence concatenation across prediction horizons.}
  \label{fig:patchtst}
\end{figure*}

\subsubsection{Comparison with Direct Sequence Concatenation. }
Considering that RAG is most widely applied in the NLP domain, where retrieved sequences are typically concatenated directly into the model’s prompt, we first evaluate whether a similar approach is effective for time series forecasting. To assess this, we conduct experiments where retrieved reference sequences are directly concatenated before the input sequence and then fed into PatchTST for model training. Figure \ref{fig:patchtst} presents the comparison between the direct sequence concatenation approach with our proposed TS-RAG framework. It is observed that TS-RAG and TS-RAG-CM consistently achieve lower MSE compared to Direct Sequence Concatenation across different output lengths. At 96 length, TS-RAG reduces MSE by 12.1\%, while TS-RAG-CM improves it by 14.2\%. For 192 length, TS-RAG achieves a 13.8\% reduction, and TS-RAG-CM improves it by 17.6\%. At 336 length, TS-RAG lowers MSE by 14.0\%, and TS-RAG-CM further reduces it by 17.4\%. The largest improvement is observed at 720 length, where TS-RAG decreases MSE by 16.2\% and TS-RAG-CM achieves an 18.2\% reduction. These results show that TS-RAG and TS-RAG-CM improve forecasting accuracy across different horizons. These results indicate that direct sequence concatenation is not as effective in time series forecasting models as it is in NLP tasks. Unlike large language models (LLMs), which possess strong generalization and generative capabilities, time series forecasting models struggle to effectively utilize concatenated historical sequences in a meaningful way. Our findings highlight the necessity of the reference fusion framework proposed in TS-RAG. By introducing a reference fusion mechanism rather than merely concatenating them, TS-RAG and TS-RAG-CM consistently achieve lower MSE across different forecasting horizons. 

\subsubsection{Comparison across different reference numbers}
\begin{figure*}
  \centering
\includegraphics[width=0.8\textwidth]{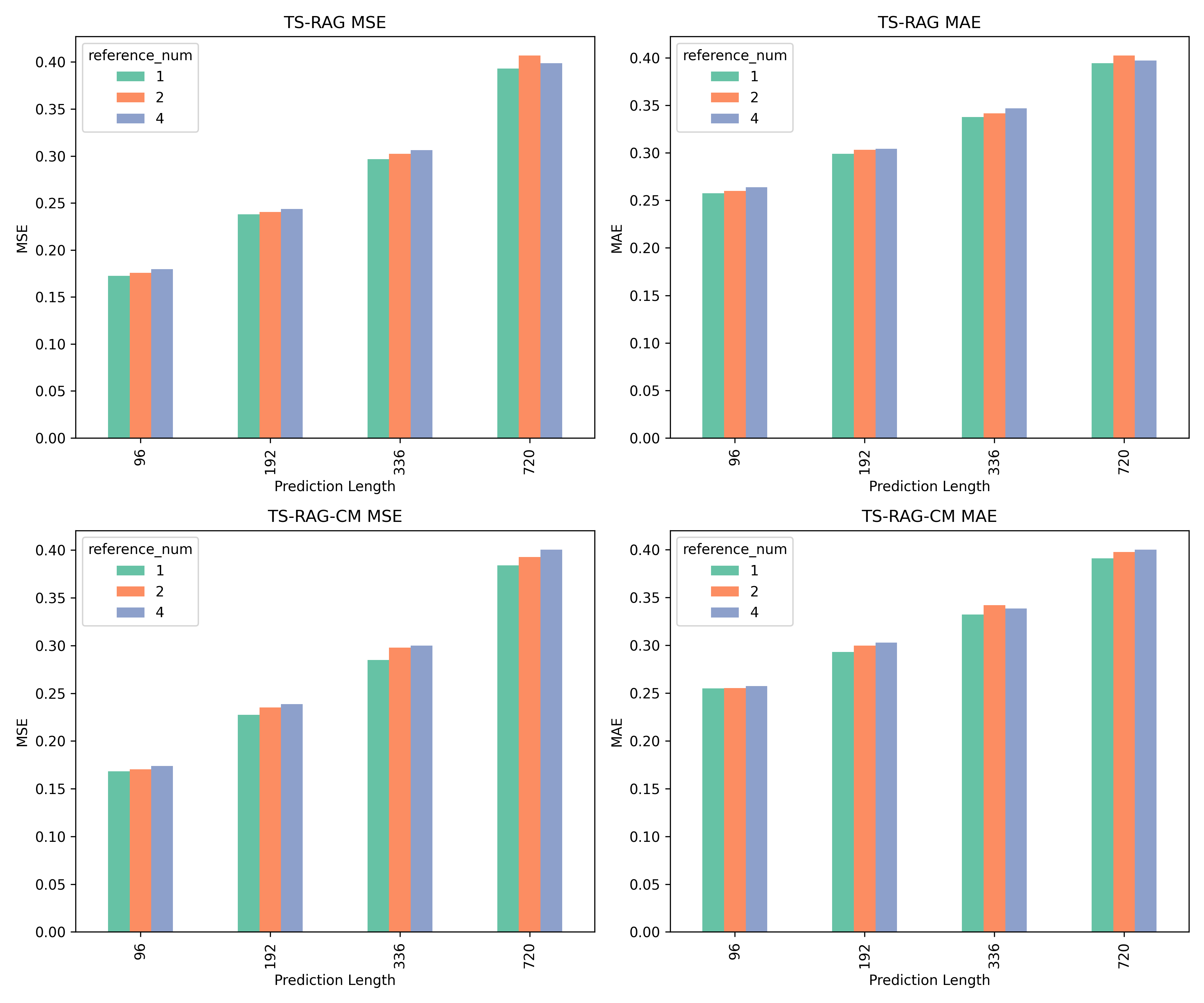}
  \caption{Comparison of TS-RAG-CM, TS-RAG across different reference numbers.}
  \Description{Forecasting error comparison for TS-RAG-CM and TS-RAG using different numbers of retrieved reference sequences.}
  \label{fig:ref_num}
\end{figure*}

Another factor that may impact model performance is the number of retrieved similar sequences. In Table \ref{tab:main_result}, we reported results considering only the top 1 similar sequence. Now, we examine how different reference numbers affect the performance of TS-RAG and TS-RAG-CM (Figure \ref{fig:ref_num}). Across different prediction lengths, the results show that TS-RAG and TS-RAG-CM achieve the best performance when using a single reference sequence. In contrast, incorporating two or four reference sequences generally leads to higher MSE and MAE, suggesting that additional retrieved sequences do not always enhance forecasting accuracy and may introduce noise. A possible explanation is that retrieving too many similar sequences may introduce redundant or conflicting patterns, making it harder for the model to extract relevant information and effectively integrate it into the forecasting process.
\subsubsection{Comparison of different retrieval approaches. }
In addition to information fusion strategies and the number of retrieved sequences, another crucial factor affecting model performance is the retrieval approach used to find similar sequences. The choice of retrieval method not only impacts forecasting accuracy but also plays a significant role in model inference efficiency. As shown in Table \ref{tab:dtw}, TCN-based retrieval achieves the best forecasting performance, with lower MSE and MAE compared to DTW-based and Euclidean Distance (ED) retrieval methods. Although TCN-based retrieval is slightly slower than ED retrieval (0.004s vs. 0.002s), it is still significantly faster than DTW-based retrieval (14.27s), making it a more practical choice for real-time applications.

\input{tables/dtw}

%% file: tables/main_result.tex
\begin{table*}[htbp]
  \caption{Multivariate forecasting results. We compare extensive competitive models under different prediction lengths. The look-back length $L$ is set to 96 for all baselines. Results are averaged from all prediction lengths $S = \{96, 192, 336, 720\}$.}
  \label{tab:main_result}
  \centering
  \resizebox{\textwidth}{!}{  
  \begin{tabular}{lcccccccccc}
    \toprule
    Model & TS-RAG-CM & TS-RAG & TimeXer & iTransformer & RLinear & PatchTST & Crossformer & TiDE & TimesNet & DLinear \\
    \midrule
    ECL & \textbf{0.169} / \underline{0.270} & 0.181 / 0.279 & \underline{0.171} / 0.270 & 0.178 / 0.270 & 0.219 / 0.298 & 0.205 / 0.290 & 0.244 / 0.334 & 0.251 / \textbf{0.244} & 0.192 / 0.295 & 0.212 / 0.300 \\
    ETTh1 & \textbf{0.437} / 0.439 & 0.438 / 0.442 & \underline{0.437} / \underline{0.437} & 0.454 / 0.447 & 0.446 / \textbf{0.434} & 0.469 / 0.454 & 0.529 / 0.522 & 0.541 / 0.507 & 0.458 / 0.450 & 0.456 / 0.452 \\
    ETTh2 & \textbf{0.366} / \underline{0.396} & 0.374 / 0.399 & \underline{0.367} / \textbf{0.396} & 0.383 / 0.407 & 0.374 / 0.398 & 0.387 / 0.407 & 0.942 / 0.684 & 0.611 / 0.550 & 0.414 / 0.427 & 0.559 / 0.515 \\
    ETTm1 & \textbf{0.379} / \textbf{0.395} & 0.383 / 0.399 & \underline{0.382} / \underline{0.397} & 0.407 / 0.410 & 0.414 / 0.407 & 0.387 / 0.400 & 0.512 / 0.496 & 0.419 / 0.419 & 0.400 / 0.406 & 0.403 / 0.407 \\
    ETTm2 & \textbf{0.266} / \textbf{0.318} & 0.275 / 0.322 & \underline{0.274} / \underline{0.322} & 0.288 / 0.332 & 0.286 / 0.327 & 0.281 / 0.326 & 0.757 / 0.610 & 0.358 / 0.404 & 0.291 / 0.333 & 0.350 / 0.401 \\
    Weather & \textbf{0.241} / \underline{0.272} & 0.252 / 0.276 & \underline{0.241} / \textbf{0.271} & 0.258 / 0.278 & 0.272 / 0.291 & 0.259 / 0.281 & 0.259 / 0.315 & 0.271 / 0.320 & 0.259 / 0.287 & 0.265 / 0.317 \\
    AVG & \textbf{0.310} / \textbf{0.348} & 0.317 / 0.353 & \underline{0.312} / \underline{0.349} & 0.328 / 0.357 & 0.335 / 0.359 & 0.331 / 0.360 & 0.540 / 0.493 & 0.409 / 0.407 & 0.336 / 0.366 & 0.374 / 0.399 \\

    \bottomrule
  \end{tabular}
  }
\end{table*}

%% file: tables/dtw.tex
\begin{table*}[htbp]
  \caption{Comparison of model performance with different retrieval approaches. The look-back length $L$ is set to 96 for all baselines. Results are averaged from all prediction lengths $S = \{96, 192, 336, 720\}$.}
  \label{tab:dtw}
  \centering
  \resizebox{0.5\textwidth}{!}{ 
  \begin{tabular}{lccc}
    \toprule
    Model & TCN & DTW & ED \\
    \midrule
    TS-RAG & 0.275 / 0.322 & 0.278 / 0.325 & 0.275 / 0.323 \\
    TS-RAG-CM & 0.266 / 0.318 & 0.271 / 0.322 & 0.270 / 0.319 \\
    \midrule
    Retrieval Time & 0.004 & 14.27 & 0.002 \\
    \bottomrule
  \end{tabular}
  }
\end{table*}

%% file: section/conclusion.tex
\section{Conclusion}
In this work, we introduced TS-RAG, a novel retrieval-augmented generation (RAG) framework for time series forecasting, which effectively integrates retrieved historical sequences into the forecasting process. Our approach addresses key limitations of existing deep learning models by incorporating external information through reference tokens, facilitating a more structured fusion of complex temporal dependencies. By leveraging efficient vector search techniques instead of computationally expensive methods like Dynamic Time Warping (DTW), TS-RAG achieves faster inference while maintaining high predictive accuracy.Our extensive evaluation on multiple real-world time series forecasting benchmarks demonstrates that TS-RAG consistently outperforms state-of-the-art models, highlighting its potential to enhance forecasting performance in both common and rare-event scenarios. Unlike standard transformer-based architectures that rely solely on internal feature extraction, our framework retrieves and utilizes relevant external sequences, allowing the model to make better-informed predictions, particularly when training data is scarce or lacks diversity.This work bridges the gap between retrieval-augmented methods and time series forecasting, showcasing how ideas from natural language processing (NLP) can be effectively adapted to the time series domain. 